\documentclass[letterpaper, 10 pt, conference]{ieeeconf}  

\IEEEoverridecommandlockouts                              

\usepackage{graphics} 
\usepackage{epsfig} 
\usepackage{times} 
\usepackage{amsmath} 
\usepackage{amssymb}  

\usepackage{amsthm}

\usepackage{multicol}
\usepackage{multirow}
\usepackage[ruled]{algorithm2e}

\usepackage{textcomp}
\usepackage{hyperref}
\usepackage{cleveref}

\usepackage[utf8]{inputenc}
\usepackage[T1]{fontenc}
\usepackage{xcolor}
\usepackage{float}
\usepackage{gensymb}
\usepackage{bm}
\usepackage{autobreak}
\usepackage{svg}
\usepackage{mathrsfs}
\usepackage{tabularx}
\usepackage{array, makecell}
\usepackage{graphicx}
\usepackage{cellspace}

\usepackage{comment}

\usepackage{soul}

\title{\LARGE \bf
MR-conditional Robotic Actuation of Concentric Tendon-Driven Cardiac Catheters
}

\author{Yifan Wang$^{1*}$, Zheng Qiu$^{1*}$, Junichi Tokuda$^2$, Ehud J. Schmidt$^3$, Aravindan Kolandaivelu$^3$, and Yue Chen$^{4}$
\thanks{Research reported in this publication is supported by the National Institute of Biomedical Imaging And Bioengineering of the National Institutes of Health under Award Number R01EB034359. The content is solely the responsibility of the authors and does not necessarily represent the official views of the National Institutes of Health.  Corresponding author: Yue Chen. }
\thanks{$^{1}$Yifan Wang and Zheng Qiu are with the Department of Mechanical Engineering, 
        Georgia Institute of Technology, Atlanta, GA 30332, USA
        {\tt\small \{wangyf, zqiu67\}@gatech.edu}}%
\thanks{$^2$Junichi Tokuda is with the Department of Radiology, Harvard Medical
School, Boston, MA 02115 USA. {\tt\small tokuda@bwh.harvard.edu}}
\thanks{$^3$Ehud J. Schmidt and Aravindan Kolandaivelu are with the Department
of Medicine, Johns Hopkins
University, Baltimore, MD 21205 USA {\tt\small eschmi17@jhu.edu, akoland1@jhmi.edu}}
\thanks{$^{4}$Yue Chen is with the Department of Biomedical Engineering, 
        Georgia Institute of Technology/Emory University, Atlanta, GA 30332, USA
        {\tt\small yue.chen@bme.gatech.edu}}%
\thanks{$^*$Yifan Wang and Zheng Qiu contributed equally to this paper.}
}

\begin{document}

\maketitle
\thispagestyle{plain}
\pagestyle{plain}

\begin{abstract}

Atrial fibrillation (AF) and ventricular tachycardia (VT) are two of the sustained arrhythmias that significantly affect the quality of life of patients. Treatment of AF and VT often requires radiofrequency ablation of heart tissues using an ablation catheter. Recent progress in ablation therapy leverages magnetic resonance imaging (MRI) for higher contrast visual feedback, and additionally utilizes a guiding sheath with an actively deflectable tip to improve the dexterity of the catheter inside the heart. This paper presents the design and validation of an MR-conditional robotic module for automated actuation of both the ablation catheter and the sheath. The robotic module features a compact design for improved accessibility inside the MR scanner bore and is driven by piezoelectric motors to ensure MR-conditionality. The combined catheter-sheath mechanism is essentially a concentric tendon-driven continuum robot and its kinematics is modeled by the constant curvature model for closed-loop position control. Path following experiments were conducted to validate the actuation module and control scheme, achieving < 2 mm average tip position error.

\end{abstract}

\section{INTRODUCTION}

Atrial fibrillation (AF) and ventricular tachycardia (VT) are arrhythmias driven by irregular and complex electrical activity propagations in the heart and can lead to several life-threatening complications, including stroke, cardiomyopathy, and heart failure. One of the most effective treatments for arrhythmia is radiofrequency ablation (RFA) of cardiac tissues to create necrotic (scar) tissue that blocks the transmission of aberrant electrical signals. This is performed by manipulating an electrophysiology (EP) catheter inside the left atrium (LA) to create a continuous circumferential ablation zone around the pulmonary veins. 

\begin{figure}[t!]
    \centering
    \vspace{2.8mm}
    \includegraphics[width = 0.95\linewidth]{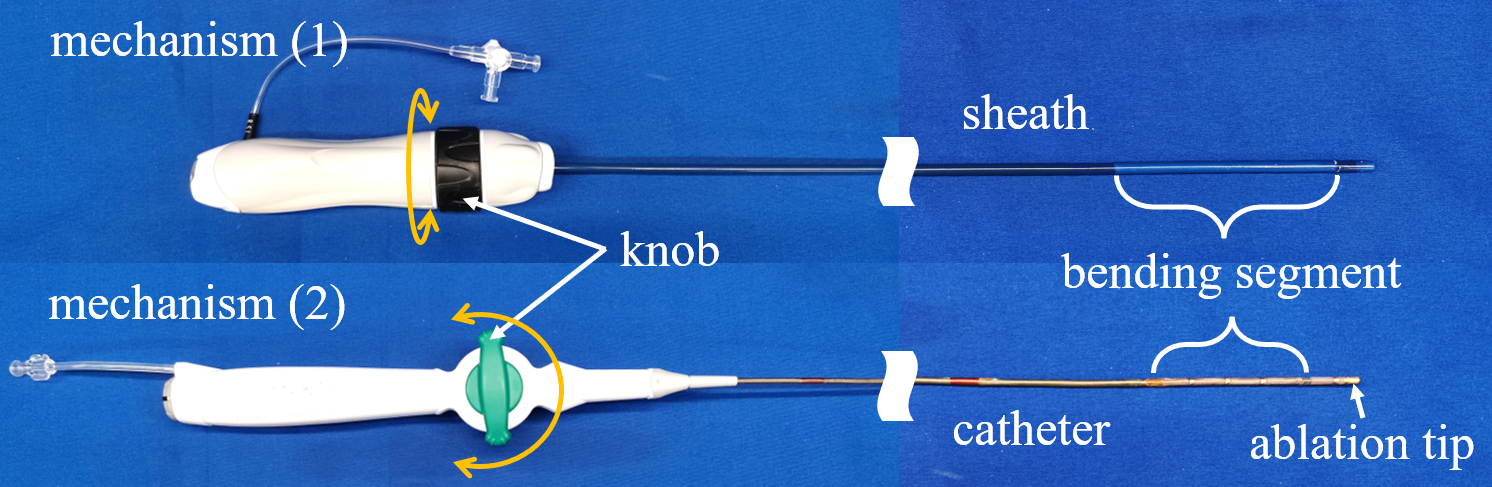}
    \vspace{-3mm}
    \caption{The actively deflectable sheath (top) and MR-tracked RF ablation catheter (bottom) used for cardiac EP procedures.} 
    \vspace{-3mm}
    \label{fig.catheter}
\end{figure}

\begin{figure*}[ht!]
    \centering
    \vspace{2.8mm}
    \includegraphics[width = 0.95\linewidth]{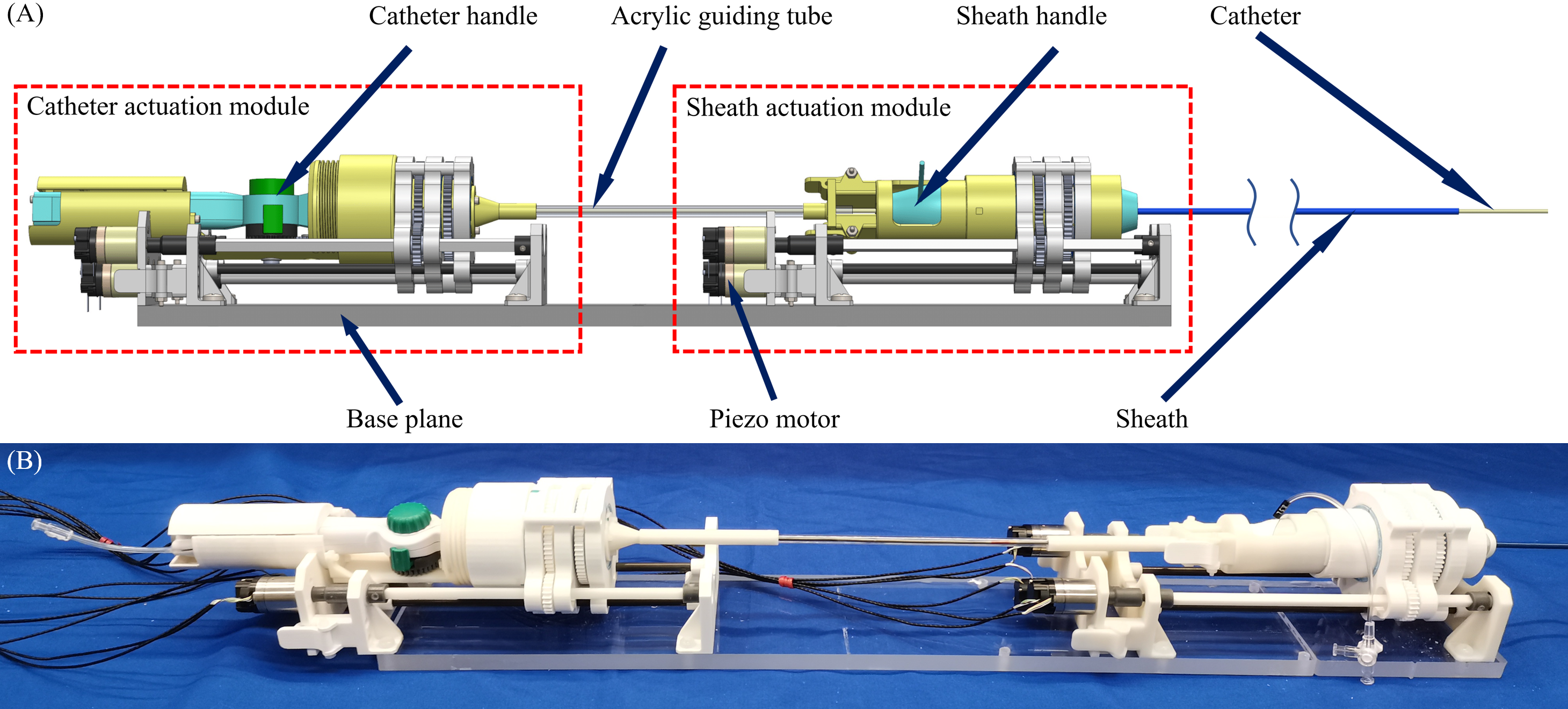}
    \vspace{-3mm}
    \caption{Overview of the proposed MR-conditional robotic platform. (A) CAD design where the gray parts are transmission modules and the yellow parts are adaptation modules. (B) Hardware prototype.} 
    \label{fig.overview}
\end{figure*}

Despite being the most effective treatment available, the outcome of conventional RFA is still suboptimal, remaining only 60-70\% effective for paroxysmal AF patients \cite{kearney2014review} and $\sim$50\% for non-paroxysmal AF patients \cite{chang2013review}. This is due to the limited ability to manipulate the catheter in a manner that reaches all desired ablation targets while also maintaining proper catheter-tissue contact force for effective ablation energy delivery \cite{chen2015intra}. There is also a lack of effective navigation methods that accurately identify sites of incomplete ablation and guide the catheter to complete ablation, resulting in the inability to create ablation lesions that are continuous, permanent, and accurately placed.

Recent progress has focused on using magnetic resonance imaging (MRI) to guide the EP procedures for improved RFA outcomes \cite{schmidt2018mri}, due to its ability to visualize fibrotic regions in the heart which can serve as substrates for arrhythmia \cite{guttman2020acute}. Based on MR images, physicians can have a better evaluation of the critical regions in the atrium and improve preoperative planning. During the procedure, MRI enables navigation of the catheter inside the heart and monitors the delivery of thermal energy, which helps to ensure that sufficient energy is delivered to achieve effective ablation while preventing damage to surrounding tissues. After the procedure, MRI can evaluate the extent of the ablated lesions and help measure the outcome of the procedure. Furthermore, MR-conditional catheters with embedded tracking coils have been developed that leverage a dedicated MR localization sequence for real-time position feedback \cite{chen_closed_2019,alipour2020MRcatheter}.

Apart from the imaging modality, the instrumentation can also be improved. Most RFA catheters are flexible tubes driven by a pair of tendons and possess 3 degrees of freedom (DoFs) actuation 
 during the surgery, including bending of the distal end, axial rotation, and translation along the catheter. While the tool flexibility significantly facilitates the manipulation inside confined environments, the lack of actuation DoFs limits the capability to enable effective position and orientation control of the catheter tip. To address this issue, some guiding sheath products have been developed recently which are equipped with similar tendon-driven mechanisms to achieve active bending at the distal section (Fig. \ref{fig.catheter}). These guiding sheaths can replace the conventional sheaths, which cannot be deflected at their tip and therefore are primarily only useful to restrict the catheter's shaft's motion, and effectively add another 3 DoFs to the catheterization toolset, thus improving the surgical outcomes \cite{Jin2023sheathreview}. The combined catheter-sheath mechanism bears similarities to the concentric tube continuum robots, but at the same time preserves the active bending capability of the tendon-driven continuum robots.

The additional DoFs, however, render the simultaneous manual control of both the catheter and sheath difficult, thus motivating a robotic solution. Robotic catheterization platforms have been developed in research labs \cite{ganji_robot-assisted_2009,cercenelli_cathrob_2017,xiang2019guidewireandcatheter,ghazbi2021autocatheter,alahmad2023forcecontrol}, while commercialized products have also emerged including the Hansen Sensei X2 system which drives the concentric tendon-driven mechanism. However, these systems are not MR-conditional and most of them can only actuate a single device (i.e. the catheter or the sheath). Despite the rapid development of MR-conditional medical robots \cite{farooq2023MRrobotreview}, only a few MR-conditional systems for clinically used cardiac RFA interventional devices exist \cite{salimi2012robocath,tavallaei2013MRcatheternav,lee2018MRcatheter}. Moreover, they are only capable of actuating a single catheter, and have relatively large sizes that limit the accessibility of the confined MR scanner bore.

In this paper, we present an MR-conditional robotic platform that is capable of simultaneously actuating an active catheter and an active sheath, both of which are clinical-grade MR-conditional products and have different driving mechanisms in the handles. 
Similar to the MR active tracking coils developed in our prior work (0.6mm resolution, 30Hz update rate, \cite{gunderman2021mr}), we implemented closed-loop position control using an adaptive constant curvature model together with position and orientation feedback from electromagnetic tracking coils. 
Bench-top experiments were conducted and the robot achieved $< 2$ mm average tip position error during path following.
This rest of the paper is organized as follows: Sec. \ref{sec_design} presents the design and fabrication of the robotic actuation module. A kinematics model of the concentric tendon-driven robot is presented in Sec. \ref{sec_model}. Sec. \ref{sec_results} provides experimental results, followed by the conclusion in Sec. \ref{sec_conclusions}. 


\section{Robot Design and Implementation} \label{sec_design}

\subsection{Clinical requirements}
The catheter and sheath both need 3 actuation DoFs including bending, axial rotation and translation. This is achieved by manipulating their handles which are typically 15-20 cm long and have $<$ 4 cm diameter. As shown in Fig. \ref{fig.catheter}, the handles have two different types of mechanisms for controlling the bending: (1) by rotating the knob (fixed to a lead screw nut) that is co-axial with the handle, or (2) by rotating the knob (fixed to a pulley) whose axis is perpendicular to the handle axis. This requires two different actuating mechanisms in the robotic platform. Additionally, to avoid the instability of long work trips, the sheath will first be fed into the heart through a blood vessel by the physician at the beginning of the procedure, and then mounted to the robot. Therefore, the robot should have easy-to-access fixtures for the handles. Lastly, MR-conditionality is indispensable, which typically requires that the robot does not contain ferromagnetic materials.

\subsection{Mechanical Design Overview}
The robot system consists of two actuation modules which are fixed to an Acrylic base plate and connected by guiding tubes, as shown in Fig. \ref{fig.overview}. Each actuation module can be divided into a transmission module and a handle adaptation module. 
The transmission module is the same for both systems while the handle adaptation module is specific to the actuating mechanism and geometry of each handle, thus facilitating modularity. The robot features a compact design with a $108$ mm $\times100$ mm $\times865$ mm overall size, thus minimizing the occupation of the patient bed or room inside the MR scanner bore. All custom-designed parts are 3D-printed either with acrylonitrile butadiene styrene (ABS) using a Stratasys F170 printer (Stratasys, MN, USA) or with Tough 2000 resin using a FormLabs Form 3B printer (FormLabs, MA, USA), depending on accuracy and mechanical strength requirements. All fasteners are made of Nylon.

\subsection{Transmission Module}
\begin{figure}[t!]
    \centering
    \vspace{2.8mm}
    \includegraphics[width = 0.95\linewidth]{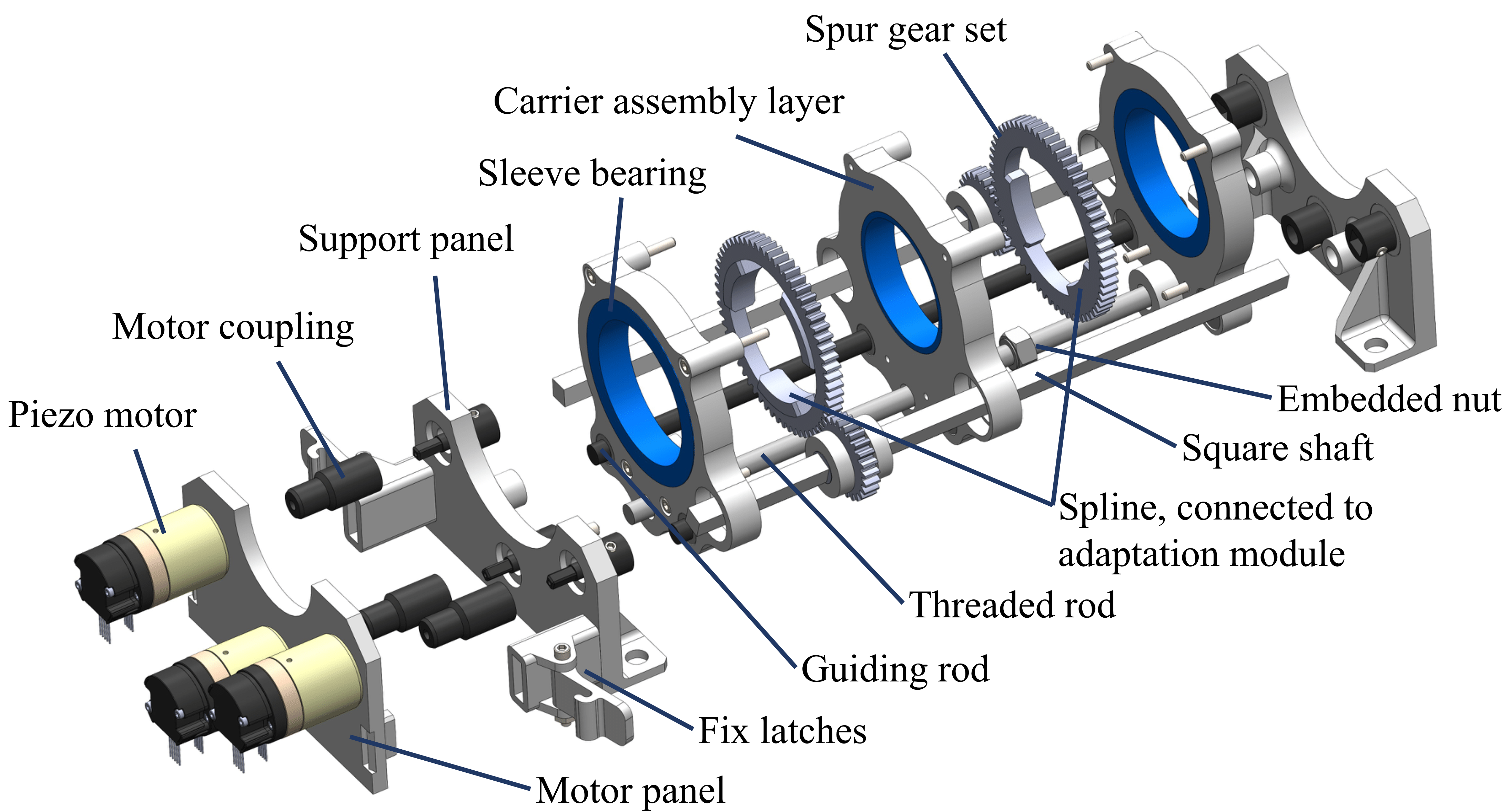}
    \vspace{-3mm}
    \caption{Design of the transmission module.} 
    \label{fig.transmission}
\end{figure}

The transmission module takes the motor outputs and transmits the motions to achieve axial rotation and translation of the handle, and also gives proper inputs to the handle adaptation module. The axial translation with 50 mm motion range is generated by a lead screw assembly, with a Nylon threaded rod and two guiding rods supported on both ends by supporting panels, as shown in Fig. \ref{fig.transmission}. The lead screw drives a carrier assembly which has an embedded nut and sleeve bearings for smooth translation along the guiding rods. The carrier assembly has three layers housing two sets of 3D-printed spur gears with a 5:2 gear ratio for axial rotation and handle input, respectively. The larger gears are supported by sleeve bearings and the smaller gears by ball bearings. The smaller gears are driven by square shafts such that they can slide on the shaft while being rotated. The square shafts and the lead screw are supported by plastic ball bearings (B688B3G, Igus, Germany) with glass balls on both ends. All shafts and rods are arranged below the handle adaptation module to make space for easy installation of the handles to the robot.

The square shafts and the lead screw are driven by non-magnetic piezo motors (LR23-50, PiezoMotor, Sweden) to ensure MR-conditionality. The motors are fixed to the motor panel, which can be quickly attached or detached from the transmission module through a pair of latches. The motor control is closed-loop based on the built-in encoders.

\subsection{Handle Adaptation Module}
The handle adaptation modules are designed to hold specific handles and transmit the rotation of the carrier gear to the tendon-driving mechanism on the handle. For this prototype, we designed adaptation modules for a sheath with a handle of mechanism type (1) and a catheter with a handle of mechanism type (2) mentioned in Sec. \ref{sec_design}. A. 

\begin{figure}[t!]
    \centering
    \vspace{2.8mm}
    \includegraphics[width = 0.95\linewidth]{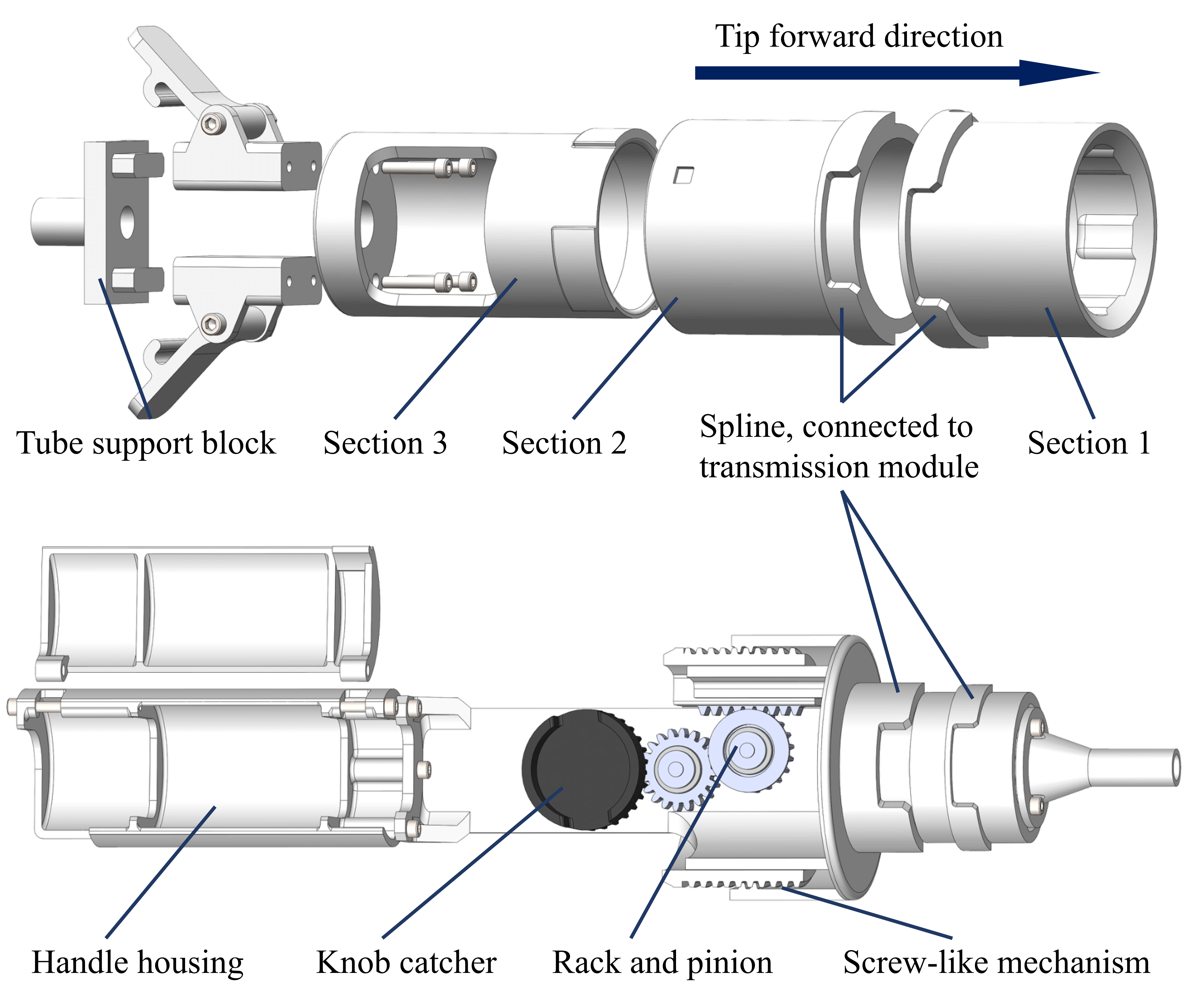}
    \vspace{-3mm}
    \caption{Design of the adaptation modules for type (1) mechanism (top) and type (2) mechanism (bottom).} 
    \label{fig.adaptation}
\end{figure}

The adaptation module for the sheath handle consists of Four sections, as shown in Fig. \ref{fig.adaptation}. Section 1 has extrusions on the inside surface that match the grooves on the surface of the sheath handle, and is driven by the first large gear in the carrier through splines to achieve axial rotation of the handle. Section 2 is similar to section 1 but matches with the nut on the handle, and is driven by the second large gear in the carrier to achieve active bending of the sheath. The axial rotation and active bending are decoupled by using the difference in the motor angles to control the rotation of the knob relative to the handle body. Section 3 is a detachable cap that extends behind the carrier to hold the handle and allows the installation of the sheath handle from the back. Section 3 is also equipped with a pair of latches used to quickly attach the guiding tubes. The guiding tubes are used to prevent buckling of the catheter between two actuation modules, and are made of telescopic Acrylic tubes to achieve extension and retraction.

The adaptation module for the catheter handle consists mainly of a handle housing. The housing is directly driven by the first large gear in the carrier for axial rotation. The cap on the housing is closed and latched with the housing to quickly fix the handle. A knob catcher is supported by the housing and matches the knob on the handle for rotating the knob. To turn the axial rotation of the carrier gear into the rotation of the knob perpendicular to the axis, we employed a lead screw-like mechanism cascaded with a rack and pinion. The second large gear in the carrier drives a housing with an internal ACME thread, which acts as a "lead screw". This then drives a nut with an external thread and slides on the outer surface of the handle housing. This lead screw-like mechanism generates an axial translation and provides self-locking. A linear gear (rack) is fixed on the inside of the nut and meshes with a circular gear (pinion) supported by the handle housing, which turns the translation into the rotation perpendicular to the handle axis. This rotation is transmitted through another gear that meshes with the gear on the knob catcher. This cascaded mechanism avoids driving the knob directly with the motor, which necessarily requires the motor itself to rotate with the handle housing as seen in \cite{tavallaei2013MRcatheternav,lee2018MRcatheter}, thus achieving a compact and static package of the actuation module. It also provides sufficient gear reduction to generate enough torque for the knob. 
 
\section{Robot Modelling and Control} \label{sec_model}
The combination of the tendon-driven catheter and sheath effectively forms a concentric tube robot with each tube actively driven by tendons. While a geometrically exact model for such a robot based on Cosserat rod theory was proposed in \cite{Chitalia23tendonCTR, xiao2023kinematics}, we note that its implementation on off-the-shelf MR-conditional catheter products is difficult since (1) the tendon actuation force or displacement required by the model is hard to measure inside a commercialized catheter handle, (2) the material properties and geometries are hard to identify due to complicated composite structure and embedded electronics (tracking coils, NavX rings, ablation probe, irrigation channel,  \cite{alipour2020MRcatheter}), and (3) frictions on tendons and backlashes of driving mechanisms in the handles are hard to incorporate in the model. Machine learning-based methods have been employed to characterize such systems with significant uncertainties in previous works \cite{kuntz2020learning,dong2022MRcathetertrack}. However, these methods require collecting large amounts of data for training, which is impractical since the data collection interferes with the clinical workflow and inevitably degrades the mechanical performance of the products. To overcome these problems, we propose to use the constant curvature model \cite{constant_curvature} that adapts online to the actual robot shape for closed-loop control. We chose the constant curvature model due to its simplicity of parameter identification and computational efficiency for online adaptation.

\subsection{Constant Curvature Model}
\begin{figure}[t!]
    \centering
    \vspace{2.8mm}
    \includegraphics[width = 0.95\linewidth]{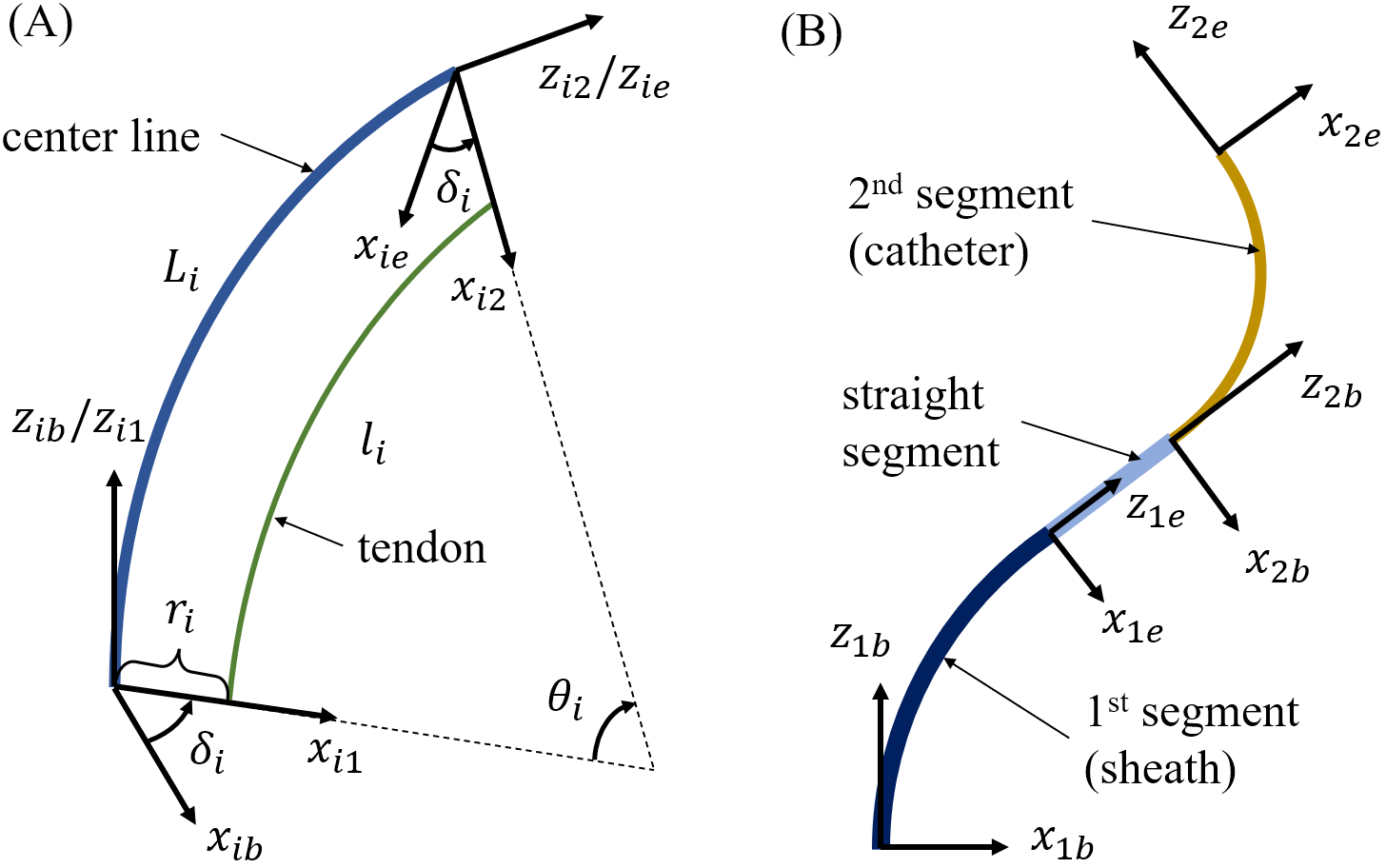}
    \vspace{-3mm}
    \caption{Illustration of the constant curvature model. (A) Kinematics of a single bending segment. (B) Kinematics of the entire robot.} 
    \label{fig.kinematics}
\end{figure}

As shown in Fig. \ref{fig.kinematics}, we place the origin of the reference frame at the end of the robot guiding tube, with the z-axis aligned with the tube. The robot centerline is represented by two circular arcs that are tangent to a straight line segment connecting them.
Define the vector of configuration variables that describes the robot shape as $\mathbf{\Psi} = [\theta_1~L_1~\delta_1~\theta_2~L_2~\delta_2]^T$, where $\theta_i$ is the bending angle of the $i$th bending segment, $L_i$ is the length of the arc, and $\delta_i$ is the angle of bending direction. To describe the kinematics of a single bending segment, we define the following frames: (1) base frame $\{ib\}$ is located at the center of the base of the $i$th segment with its z-axis perpendicular to the cross-section; (2) bending plane frame $\{i1\}$ is fixed on the base cross-section of the $i$th segment such that its x-axis is aligned with the tendons of the $i$th segment and it can be obtained by rotating $\{ib\}$ around the z-axis by $\delta_i$; (3) bending plane frame $\{i2\}$ is fixed on the end cross-section of the $i$th segment such that its x-axis is aligned with the tendons of the $i$th segment and its orientation can be obtained by rotating $\{i1\}$ around its y-axis by $\theta_i$; (4) end frame $\{ie\}$ is located at the center of the end of the $i$th segment with its z-axis perpendicular to the cross-section such that it can be obtained by rotating $\{i2\}$ around the z-axis by $-\delta_i$. Using the geometry of the arc, we obtain the following transformation:
\begin{equation}\label{eq:segment_kinematics}
\begin{aligned}
    ^{ib}\mathbf{R}_{ie} &= \mathbf{R}_z(\delta_i)\mathbf{R}_y(\theta_i)\mathbf{R}_z(-\delta_i)\\
    ^{ib}\mathbf{p}_{ie} &= \frac{L_i}{\theta_i}\begin{bmatrix}
        (1-\cos\theta_i)\cos\delta_i \\ (1-\cos\theta_i)\sin\delta_i \\ \sin\theta_i
    \end{bmatrix}
\end{aligned}
\end{equation}
where $^{ib}\mathbf{R}_{ie} \in SO(3)$ is the rotation matrix from $\{ib\}$ to $\{ie\}$, $^{ib}\mathbf{p}_{ie}$ is the position coordinates of $\{ie\}$ in $\{ib\}$, and $\mathbf{R}_z(\delta)$ is the rotation around z-axis by $\delta$. It is then straightforward to obtain the kinematics of the entire robot:
\begin{equation}
    ^{1b}\mathbf{T}_{2e}(\mathbf{\Psi}) = ^{1b}\mathbf{T}_{1e}\mathbf{T}_{z}(L_n)^{2b}\mathbf{T}_{2e}
\end{equation}
where $^{ib}\mathbf{T}_{ie} \in SE(3)$ is the homogeneous transformation matrix from $\{ib\}$ to $\{ie\}$, and $\mathbf{T}_{z}(L_n)$ is the linear translation along z-axis for $L_n$, which is the length of the straight middle segment.

The robot Jacobian of the $i$th segment w.r.t. $\mathbf{\Psi}$ can be obtained by taking time derivatives of (\ref{eq:segment_kinematics}) and noting that the angular velocity is
\begin{equation}
    ^{ib}\mathbf{\omega}_{ie} = \Dot{\delta_i}\mathbf{e}_z + ^{ib}\mathbf{R}_{i1}(\Dot{\theta_i}\mathbf{e}_y - ^{i1}\mathbf{R}_{i2}\Dot{\delta_i}\mathbf{e}_z)
\end{equation}
where $\mathbf{e}_y$ and $\mathbf{e}_z$ are unit basis vectors. This gives the linear and angular Jacobians as follows:
\begin{equation}\label{eq:segment_Jacobians}
\begin{aligned}
    \mathbf{J}_{iv} &= \begin{bmatrix}
        c\delta_i L_i(c\theta_i-h_i)/\theta_i & c\delta_ih_i & -L_is\delta_ih_i\\
        s\delta_iL_i(s\theta_i-h_i)/\theta_i & s\delta_ih_i & L_ic\delta_ih_i\\
        L_i(c\theta_i-s\theta_i/\theta_i)/\theta_i & s\theta_i/\theta_i & 0
    \end{bmatrix}\\
    \mathbf{J}_{i\omega} &= \begin{bmatrix}
        -s\delta_i & 0 & -c\delta_is\theta_i\\
        c\delta_i & 0 & -s\delta_is\theta_i\\
        0 & 0 & 1-c\theta_i
    \end{bmatrix}
\end{aligned}
\end{equation}
where $c$ and $s$ denote sin and cos, respectively, and $h_i = (1-c\theta_i)/\theta_i$. The linear Jacobian of the entire robot w.r.t. $\mathbf{\Psi}$ can then be calculated as:
\begin{equation}
    \mathbf{J}_v = \begin{bmatrix}
        \mathbf{J}_{1v}-~^{1e}\Hat{\mathbf{p}}_{2e}\mathbf{J}_{1w} & ^{1b}\mathbf{R}_{2b}\mathbf{J}_{2v}
    \end{bmatrix}
\end{equation}
such that $^{1b}\Dot{\mathbf{p}}_{2e} = \mathbf{J}_v\Dot{\mathbf{\Psi}}$, where $\Hat{\mathbf{p}}$ is the skew-symmetric matrix of $\mathbf{p}$.

We next describe the robot actuation. Define robot actuation variables $\mathbf{q} = [\delta_1~\beta_1~\gamma_1~\delta_2~\beta_2~\gamma_2]^T$, where $\delta$ is the axial rotation angle corresponding to the previously defined configuration variables, $\beta$ is the axial translation distance, $\gamma$ is the rotation angle of the handle nob, and subscripts 1 and 2 denote the sheath and catheter, respectively. From the constant curvature geometry, we have
\begin{equation}
    \begin{aligned}
        L_i - l_i &= r_i\theta_i\\
        L_1 - l_{21} &= r_2\theta_1\cos(\delta_1 - \delta_2)
    \end{aligned}
\end{equation}
where $l_{21}$ is the length of the tendon of the catheter that passes through the sheath, and the second equation gives the catheter tendon displacement caused by the coupled bending in the 1st segment.

Mechanically, the tendon displacement is linear w.r.t. the rotation angle of the knob for both the sheath's and catheter's handles. We thus assume the following relationships between $\mathbf{q}$ and $\mathbf{\Psi}$:
\begin{equation}\label{eq:actuation_to_shape}
\begin{aligned}
    \theta_1 &= k_1\gamma_1\\
    \theta_2 &= k_2\gamma_2 + k_c\theta_1\cos(\delta1 - \delta2)\\
    L_i &= \beta_i + b_i
\end{aligned}
\end{equation}
where $k_i$, $k_c$, and $b_i$ are parameters that can be easily identified from the robot. From the above relationship, it is straightforward to calculate the Jacobian $\mathbf{J}_q$ such that $\Dot{\mathbf{\Psi}} = \mathbf{J}_q\Dot{\mathbf{q}}$. We finally obtain the robot Jacobian for closed-loop control as $\mathbf{J} = \mathbf{J}_v\mathbf{J}_q$.

\subsection{Closed-loop Control}
In clinical practice, MR-conditional catheters are often equipped with MR tracking coils that help extract the coil positions and orientations from the MR image. This can be leveraged to perform real-time MR-guided task-space closed-loop control of the robot. We use the resolved-rates algorithm for position control of the catheter tip. The actuation space update rule is as follows:
\begin{equation}
    \Delta \mathbf{q} = \alpha\mathbf{J}^{\dagger}(\mathbf{p}_{desired} - \Bar{\mathbf{p}})
\end{equation}
where $\alpha$ is the control gain, $\mathbf{J}^{\dagger}$ is the pseudo-inverse of $\mathbf{J}$, and $\Bar{\mathbf{p}}$ is the position feedback of the controlled point on the robot.

We note that the configuration calculated directly from $\mathbf{q}$ using (\ref{eq:actuation_to_shape}) is rarely accurate enough for the real robot. However, the Jacobian $\mathbf{J}_q$ still provides valid information on the instantaneous relationship between $\mathbf{\Psi}$ and $\mathbf{q}$. To obtain a $\mathbf{J}$ that is sufficient for the convergence of the closed-loop control, a $\mathbf{J}_v$ that is accurate to the robot shape is required. We hence estimate $\mathbf{\Psi}$ in every iteration of the resolved-rates by using the tracking coil feedbacks from both the catheter and the sheath and solving the following optimization problem:
\begin{equation}\label{eq:least-square}
    \min_{\mathbf{\Psi}}~\sum_{i=1}^2(w_{pi}(\Bar{\mathbf{p}}_i - \mathbf{p}_i(\mathbf{\Psi})) + w_{ti}(\Bar{\mathbf{t}}_i - \mathbf{t}_i(\mathbf{\Psi})))
\end{equation}
where $\Bar{\mathbf{p}}_i$ and $\Bar{\mathbf{t}}_i$ are tracking coil position and orientation feedback of the $i$th segment, and $\mathbf{p}_i(\mathbf{\Psi})$ and $\mathbf{t}_i(\mathbf{\Psi})$ are tracking coil position and orientation calculated from the forward kinematics. 

\section{Experimental Results} \label{sec_results}

\subsection{Experimental Setup}

\begin{figure}[t!]
    \centering
    \vspace{2.8mm}
    \includegraphics[width = 0.95\linewidth]{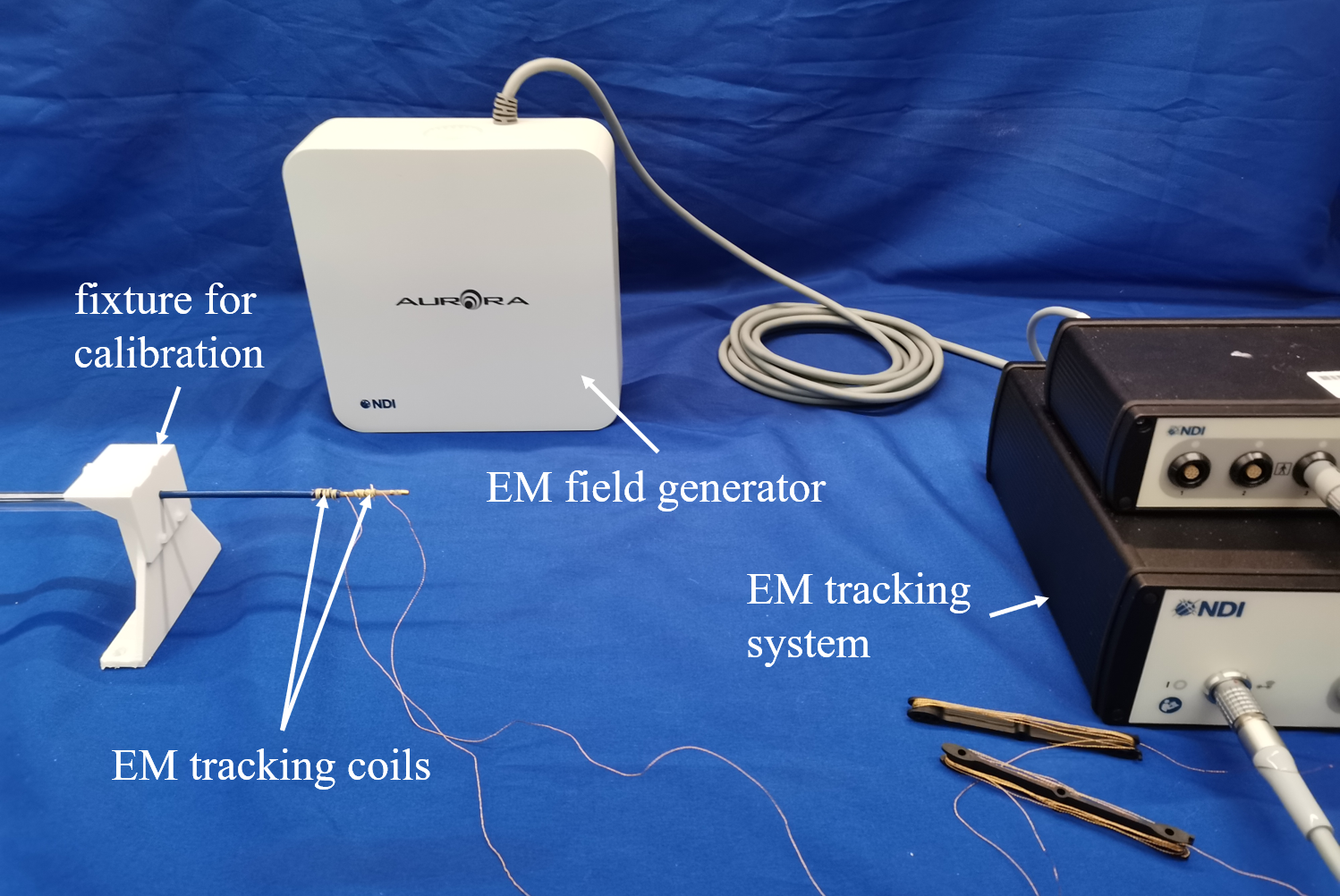}
    \vspace{-3mm}
    \caption{Experimental setup for calibration and path following.} 
    \label{fig.setup}
\end{figure}

Bench-top experiments were conducted to validate the performance of the robot hardware and control algorithm. As shown in Fig. \ref{fig.setup}, two 5-DoF electromagnetic (EM) tracking coils were mounted to the end of the bending segments of the catheter and sheath, respectively. The positions and tangent directions of the coils were measured by an EM tracking system (Aurora, NDI Medical, Ontario, Canada), emulating the clinical scenario where embedded tracking coils are tracked by the MR scanner \cite{chen_closed_2019,alipour2020MRcatheter}. The less flexible stems of the sheath and catheter were constrained in a rigid tube to be straight and only the active bending segments protruded out of the tube. A 3D-printed fixture was fixed to the exit of the tube with slots for tracking coils, allowing calibration of the transformation from the tracking system frame to the base frame of the robot. The control algorithm was implemented in Matlab App Designer which communicates with motor controllers (PMD401, PiezoMotor, Sweden) and the EM tracking system both through serial ports and allows real-time monitoring and control of the system. The actuation parameters in (\ref{eq:actuation_to_shape}) were identified by actuating the robot to several configurations, measuring the bending angles, and least-square fitting.

\subsection{Path Following}

The robot was actuated to follow a circular path with the point on the catheter where the tracking coil was mounted. This is the typical path used in RF ablation as the catheter is expected to create a circular ablation zone around the pulmonary veins (PVs). The path was evenly discretized into 72 points, and a maximum of 20 control cycles were allowed for the tracking of each target point. The threshold of convergence was set to 1 mm for the resolved-rates algorithm. For comparison, we employed three control schemes: (1) open-loop control, (2) closed-loop control without online shape fitting, and (3) closed-loop control with online shape fitting. 

\begin{figure}[t!]
    \centering
    \vspace{2.8mm}
    \includegraphics[width = 0.95\linewidth]{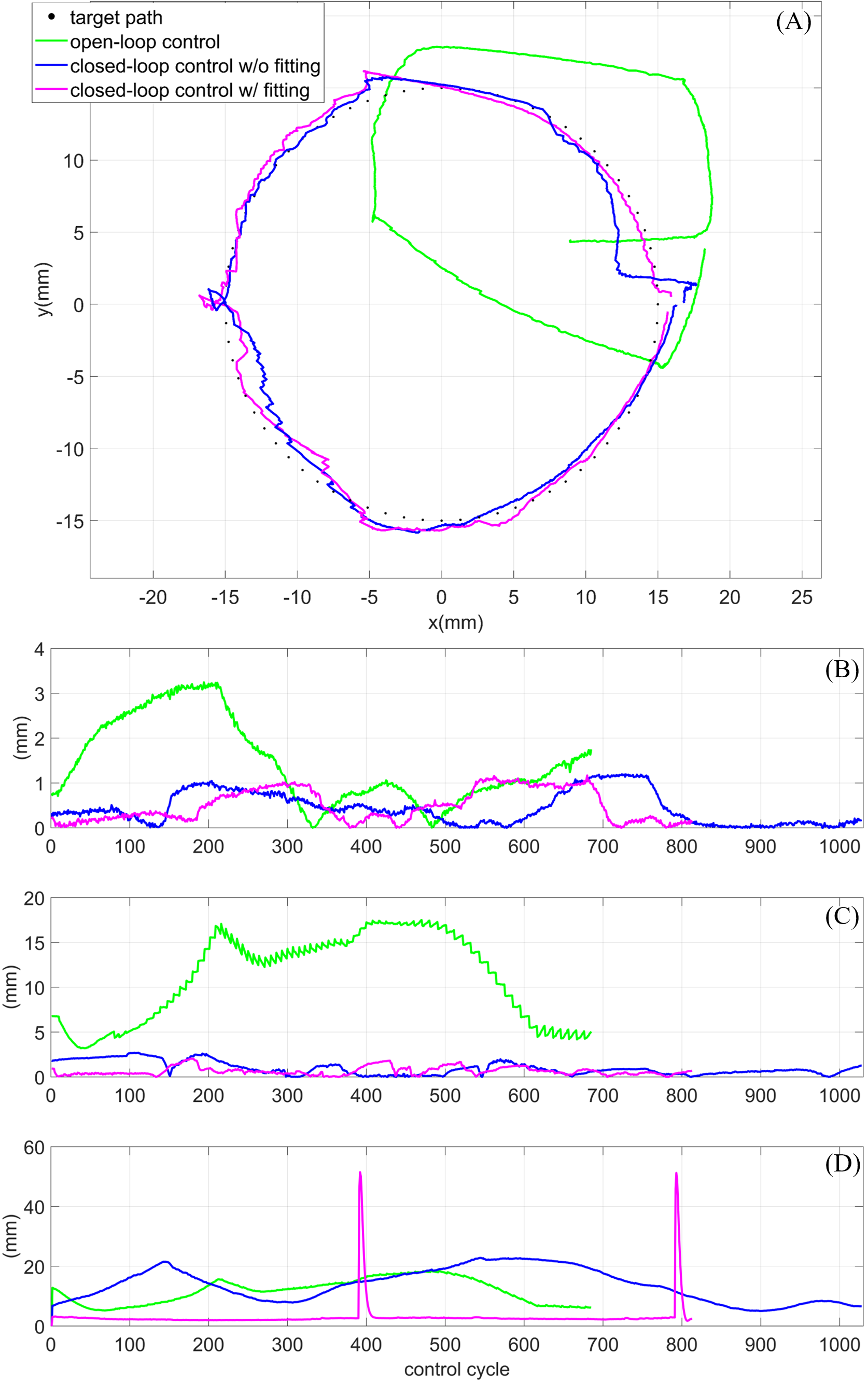}
    \vspace{-3mm}
    \caption{Experimental results. (A) Target path and actual paths measured by the tracking coil. Motion direction is counter-clockwise. (B) Position errors in the direction perpendicular to the plane of the circle path. (C) Position errors in the plane of the circle path. (D) Position error of the catheter tip between the constant curvature model and the tracking coil measurement. } 
    \label{fig.results}
\end{figure}

As shown in Fig. \ref{fig.results}, both closed-loop control schemes achieved significantly lower path-following errors than the open-loop scheme. The average position error perpendicular to the path plane is 1.45 mm for the open-loop control, 0.43 mm for the closed-loop control without shape fitting, and 0.50 mm for the closed-loop control with shape fitting, while the average in-plane position error is 11.08 mm for the open-loop control, 0.65 mm for the closed-loop control without shape fitting, and 0.98 mm for the closed-loop control with shape fitting. Since the robot length outside the guiding tube is about 125 mm, the tip position error consists of less than 4\% of the robot length in the worst case, demonstrating the possibility of controlling commercialized catheters without sophisticated modeling or extensive data collecting. While there is not an obvious difference in the path-following performance between the two closed-loop schemes, the control with shape fitting achieved 3.18 mm average tip position error between the constant curvature model and the sensor measurement, which is significantly lower than the 13.72 mm error of the control without shape fitting. This can potentially be useful for obstacle avoidance and online estimation of robot parameters for better control performance.

In this preliminary study, several limitations are identified that require future work. As shown in Fig. \ref{fig.results}-(D), sudden spikes exist in the shape fitting error of the proposed control scheme, which were caused by the failure of the optimization solver for (\ref{eq:least-square}). Also, solving nonlinear optimization online is time-consuming. A possible improvement would be using recursive nonlinear regression techniques to estimate the shape variables $\mathbf{\Psi}$. Similar methods can also be used to estimate the actuation parameters in (\ref{eq:actuation_to_shape}) which are fixed in the current study. Due to the manufacturing tolerances and complicated nonlinear mechanics of the robot system, these parameters cannot hold true for the entire robot workspace. However, using filtering methods such as the unscented Kalman filter, it is possible to estimate the nonlinear relationship between $\mathbf{q}$ and $\mathbf{\Psi}$ that works locally to improve the closed-loop control performance. Finally, we can incorporate a hysteresis model in the control scheme to overcome the backlash in the system.

\section{CONCLUSIONS} \label{sec_conclusions}
This paper presents preliminary work on the design, implementation, and control of an MR-conditional robotic actuation platform for simultaneous control of combined deflectable catheter and guiding sheath for cardiac RFA ablation procedures. The platform is completely made of plastics and actuated by non-magnetic piezo motors. The design is compact and allows for quick installation of the catheter and sheath, thus potentially reducing the interruption to the clinical workflow. This paper also demonstrates the possibility of controlling the commercial catheter and sheath using a simple constant curvature model and tracking coil feedback, which facilitates future works on more accurate adaptive control of such instruments. Our future work will focus on control improvement and performing RFA ablations in swine models of arrhythmia inside the MRI scanner.






\bibliographystyle{ieeetr}
\bibliography{references}

\begin{thebibliography}{10}

\bibitem{kearney2014review}
K.~Kearney, R.~Stephenson, K.~Phan, W.~Y. Chan, M.~Y. Huang, and T.~D. Yan, ``A systematic review of surgical ablation versus catheter ablation for atrial fibrillation,'' {\em Annals of Cardiothoracic Surgery}, vol.~3, no.~1, 2014.

\bibitem{chang2013review}
H.-Y. Chang, L.-W. Lo, Y.-J. Lin, S.-L. Chang, Y.-F. Hu, C.-H. Li, T.-F. Chao, F.-P. Chung, T.~L. Ha, R.~Singhal, E.~Chong, W.-H. Yin, H.-M. Tsao, M.-H. Hsien, and S.-A. Chen, ``Long-term outcome of catheter ablation in patients with atrial fibrillation originating from nonpulmonary vein ectopy,'' {\em Journal of Cardiovascular Electrophysiology}, vol.~24, no.~3, pp.~250--258, 2013.

\bibitem{chen2015intra}
Y.~Chen, Z.~T. Tse, W.~Wang, R.~Y. Kwong, W.~G. Stevenson, and E.~J. Schmidt, ``Intra-cardiac mr imaging \& mr-tracking catheter for improved mr-guided ep,'' {\em Journal of Cardiovascular Magnetic Resonance}, vol.~17, no.~1, pp.~1--2, 2015.

\bibitem{schmidt2018mri}
E.~J. Schmidt and H.~R. Halperin, ``{MRI} use for atrial tissue characterization in arrhythmias and for {EP} procedure guidance,'' {\em The International Journal of Cardiovascular Imaging}, vol.~34, pp.~81--95, Jan. 2018.

\bibitem{guttman2020acute}
M.~A. Guttman, S.~Tao, S.~Fink, R.~Tunin, E.~J. Schmidt, D.~A. Herzka, H.~R. Halperin, and A.~Kolandaivelu, ``Acute enhancement of necrotic radio-frequency ablation lesions in left atrium and pulmonary vein ostia in swine model with non-contrast-enhanced t1-weighted mri,'' {\em Magnetic resonance in medicine}, vol.~83, no.~4, pp.~1368--1379, 2020.

\bibitem{chen_closed_2019}
Y.~Chen, J.~Howard, I.~Godage, and S.~Sengupta, ``Closed {Loop} {Control} of an {MR}-{Conditional} {Robot} with {Wireless} {Tracking} {Coil} {Feedback},'' {\em Annals of Biomedical Engineering}, vol.~47, pp.~2322--2333, nov 2019.

\bibitem{alipour2020MRcatheter}
A.~Alipour, E.~S. Meyer, C.~L. Dumoulin, R.~D. Watkins, H.~Elahi, W.~Loew, J.~Schweitzer, G.~Olson, Y.~Chen, S.~Tao, M.~Guttman, A.~Kolandaivelu, H.~R. Halperin, and E.~J. Schmidt, ``Mri conditional actively tracked metallic electrophysiology catheters and guidewires with miniature tethered radio-frequency traps: Theory, design, and validation,'' {\em IEEE Transactions on Biomedical Engineering}, vol.~67, no.~6, pp.~1616--1627, 2020.

\bibitem{Jin2023sheathreview}
X.~Jin, Y.~Zhou, Y.~Wu, and M.~Xie, ``Safety and efficacy of steerable versus non-steerable sheaths for catheter ablation of atrial fibrillation systematic review and meta-analysis,'' {\em BMJ Open}, vol.~13, no.~9, 2023.

\bibitem{ganji_robot-assisted_2009}
Y.~Ganji, F.~Janabi-Sharifi, and A.~N. Cheema, ``Robot-assisted catheter manipulation for intracardiac navigation,'' {\em International Journal of Computer Assisted Radiology and Surgery}, vol.~4, pp.~307--315, June 2009.

\bibitem{cercenelli_cathrob_2017}
L.~Cercenelli, B.~Bortolani, and E.~Marcelli, ``{CathROB}: {A} {Highly} {Compact} and {Versatile} {Remote} {Catheter} {Navigation} {System},'' {\em Applied Bionics and Biomechanics}, vol.~2017, pp.~1--13, 2017.

\bibitem{xiang2019guidewireandcatheter}
Y.~Xiang, H.~Shen, L.~Xie, and H.~Wang, ``Master-slave guidewire and catheter robotic system for cardiovascular intervention,'' in {\em 2019 28th IEEE International Conference on Robot and Human Interactive Communication (RO-MAN)}, pp.~1--6, 2019.

\bibitem{ghazbi2021autocatheter}
S.~Norouzi-Ghazbi, A.~Mehrkish, I.~Abdulhafiz, T.~Abbasi-Hashemi, A.~Mahdi, and F.~Janabi-Sharifi, ``Design and experimental evaluation of an automated catheter operating system,'' {\em Artificial Organs}, vol.~45, no.~6, pp.~E171--E186, 2021.

\bibitem{alahmad2023forcecontrol}
O.~Al-Ahmad, M.~Ourak, J.~Vlekken, and E.~V. Poorten, ``Force control with a novel robotic catheterization system based on braided sleeve grippers,'' {\em IEEE Transactions on Medical Robotics and Bionics}, vol.~5, no.~3, pp.~602--613, 2023.

\bibitem{farooq2023MRrobotreview}
M.~U. Farooq and S.~Y. Ko, ``A decade of mri compatible robots: Systematic review,'' {\em IEEE Transactions on Robotics}, vol.~39, no.~2, pp.~862--884, 2023.

\bibitem{salimi2012robocath}
A.~Salimi, A.~Ramezanifar, J.~Mohammadpour, K.~Grigoriadis, and N.~V. Tsekos, ``{ROBOCATH: A Patient-Mounted Parallel Robot to Position and Orient Surgical Catheters},'' in {\em Dynamic Systems and Control Conference}, vol.~3, pp.~471--480, 10 2012.

\bibitem{tavallaei2013MRcatheternav}
M.~A. Tavallaei, Y.~Thakur, S.~Haider, and M.~Drangova, ``A magnetic-resonance-imaging-compatible remote catheter navigation system,'' {\em IEEE Transactions on Biomedical Engineering}, vol.~60, no.~4, pp.~899--905, 2013.

\bibitem{lee2018MRcatheter}
K.-H. Lee, K.~C.~D. Fu, Z.~Guo, Z.~Dong, M.~C.~W. Leong, C.-L. Cheung, A.~P.-W. Lee, W.~Luk, and K.-W. Kwok, ``Mr safe robotic manipulator for mri-guided intracardiac catheterization,'' {\em IEEE/ASME Transactions on Mechatronics}, vol.~23, no.~2, pp.~586--595, 2018.

\bibitem{gunderman2021mr}
A.~L. Gunderman, E.~J. Schmidt, M.~Morcos, J.~Tokuda, R.~T. Seethamraju, H.~R. Halperin, A.~N. Viswanathan, and Y.~Chen, ``Mr-tracked deflectable stylet for gynecologic brachytherapy,'' {\em IEEE/ASME Transactions on Mechatronics}, vol.~27, no.~1, pp.~407--417, 2021.

\bibitem{Chitalia23tendonCTR}
Y.~Chitalia, A.~Donder, and P.~E. Dupont, ``Modeling tendon-actuated concentric tube robots,'' in {\em 2023 International Symposium on Medical Robotics (ISMR)}, pp.~1--7, 2023.

\bibitem{xiao2023kinematics}
Q.~Xiao, M.~Musa, I.~S. Godage, H.~Su, and Y.~Chen, ``Kinematics and stiffness modeling of soft robot with a concentric backbone,'' {\em Journal of Mechanisms and Robotics}, vol.~15, no.~5, p.~051011, 2023.

\bibitem{kuntz2020learning}
A.~Kuntz, A.~Sethi, R.~J. Webster, and R.~Alterovitz, ``Learning the complete shape of concentric tube robots,'' {\em IEEE Transactions on Medical Robotics and Bionics}, vol.~2, no.~2, pp.~140--147, 2020.

\bibitem{dong2022MRcathetertrack}
Z.~Dong, X.~Wang, G.~Fang, Z.~He, J.~D.-L. Ho, C.-L. Cheung, W.~L. Tang, X.~Xie, L.~Liang, H.-C. Chang, C.~K. Ching, and K.-W. Kwok, ``Shape tracking and feedback control of cardiac catheter using {MRI}-guided robotic platform—validation with pulmonary vein isolation simulator in mri,'' {\em IEEE Transactions on Robotics}, vol.~38, no.~5, pp.~2781--2798, 2022.

\bibitem{constant_curvature}
R.~J. Webster and B.~A. Jones, ``Design and kinematic modeling of constant curvature continuum robots: A review,'' {\em The International Journal of Robotics Research}, vol.~29, no.~13, pp.~1661--1683, 2010.

\end{thebibliography}

\end{document}